\documentclass[10pt,conference]{IEEEtran}
\usepackage{graphicx}
\usepackage{mathptmx}
\usepackage{amsmath}

\begin{document}

\title{An Improved Algorithm for Eye Corner Detection}
\author{Anirban Dasgupta, Anshit Mandloi, Anjith George and Aurobinda Routray\\Department of Electrical Engineering\\
Indian Institute of Technology\\
Kharagpur, India 721302\\anirban1828@gmail.com, anmandloi@gmail.com, anjith2006@gmail.com, aurobinda.routray@gmail.com}
\maketitle
\begin{abstract}
In this paper, a modified algorithm for the detection of nasal and temporal eye corners is presented. The algorithm is a modification of the Santos and Proenka Method. In the first step, we detect the face and the eyes using classifiers based on Haar-like features. We then segment out the sclera, from the detected eye region. From the segmented sclera, we segment out an approximate eyelid contour. Eye corner candidates are obtained using Harris and Stephens corner detector. We introduce a post-pruning of the Eye corner candidates to locate the eye corners, finally. The algorithm has been tested on Yale, JAFFE databases as well as our created database.
%
%
%
%
%
%
%
%
%
\end{abstract}

\section{INTRODUCTION}
\begin{figure}[h]
\includegraphics[width=3in]{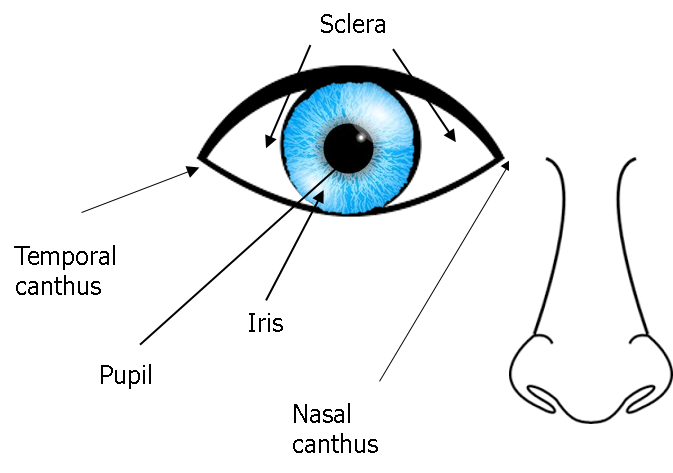}
\caption{Temporal and Nasal Canthus}
\label{canthus}
\end{figure}

Eye corners are the regions where the upper and lower eyelids meet \cite{zhu2002subpixel}. The two corners are technically named as temporal and nasal canthus as shown in Fig. \ref{canthus}. In applications, where eye movements are analyzed, eye corners serve as the reference points \cite{haiying2009novel}. The reason behind this is that the eye corners are more stable than the iris since its shape or orientation is not affected by the gaze direction or the state of eye closure.\\Eye Corner detection has been a challenging problem in computer vision owing to the following issues.
\begin{itemize}
\item eye corner in an image is not necessarily a single pixel
\item nasal eye corners can be occluded by nose
\item illumination effects and shadows may remove information related to the corner
\end{itemize}
The above issues make eye corner detection, a more challenging task than a classical corner detection in computer vision. The important works which aim at addressing the eye corner detection issue are provided in Table \ref{lit}.\\This work attempts to improve the Santos and Proenca Method \cite{santos2011robust} for eye corner detection, thereby addressing some of the above issues. The significant contributions of this work are as follows:
\begin{itemize}
\item An online framework for eye corner detection
\item Improvement of the Santos and Proenka Method for webcam quality images
\item Post pruning of eye corner candidates
\end{itemize}
\begin{table}[htp]
\centering
\caption{Earlier Works}
\label{lit}
\begin{tabular}{|l|l|}
\hline
\multicolumn{1}{|l|}{\bf Author} & \multicolumn{1}{l|}{\bf Work}                                                                         \\ \hline
Xia et al. \cite{haiying2009novel}                   & Variance Projection Functions                                                                     \\ \hline
Zhu et al.    \cite{zhu2002subpixel}              & \begin{tabular}[c]{@{}l@{}}Contour extraction and ellipse\\ fitting\end{tabular}                  \\ \hline
Xu et al.    \cite{xu2008semantic}                & Semantic Features extraction                                                                      \\ \hline
Xu et al.   \cite{xu2009real}                 & \begin{tabular}[c]{@{}l@{}}Improved Local Projection\\ Functions and Circle Integral\end{tabular} \\ \hline
Santos and Proenca \cite{santos2011robust}          & \begin{tabular}[c]{@{}l@{}}Eye contour approximation on\\ sclera segmented eye image\end{tabular} \\ \hline
\end{tabular}
\end{table}
\begin{figure}
\includegraphics[width=\linewidth]{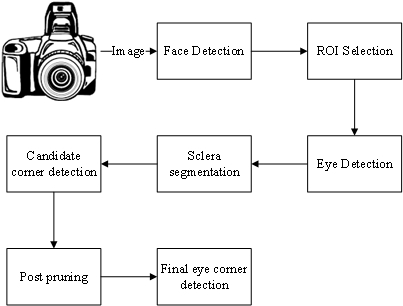}
\caption{Proposed Overall Framework}
\label{fram}
\end{figure}
\begin{figure*}[h]
\includegraphics[width=7in]{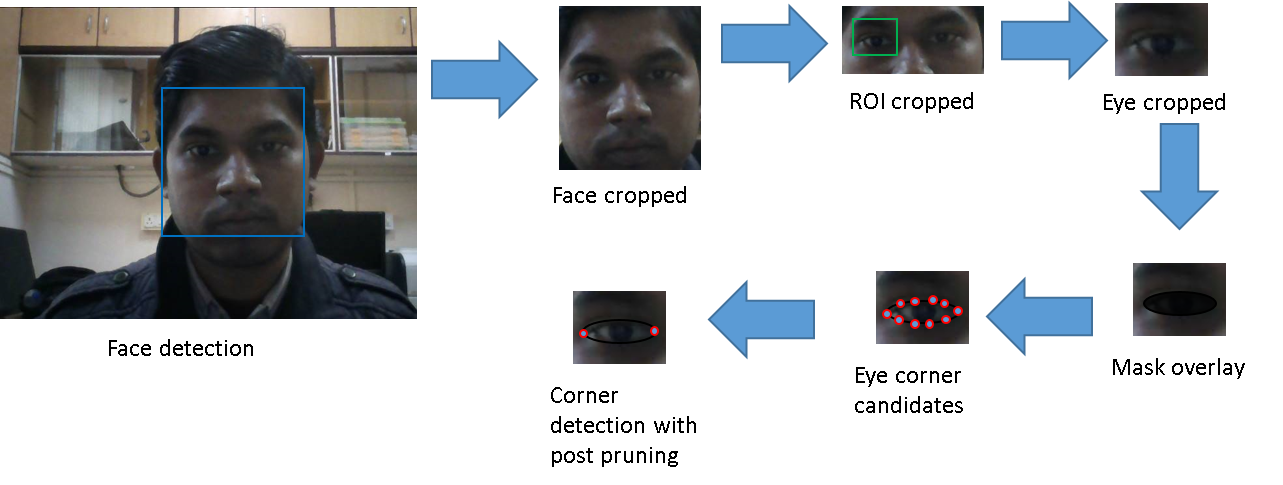}
\caption{The overall scheme}
\end{figure*}
 \begin{figure}
\includegraphics[width=3in]{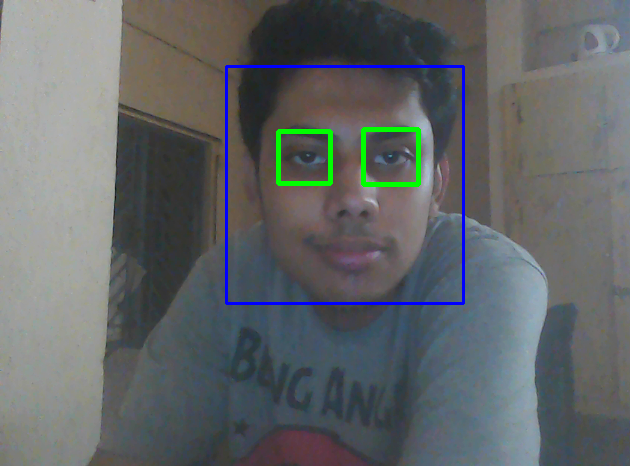}
\caption{Face and Eye Detection}
 \end{figure}
\begin{figure*}
\includegraphics[width=\textwidth]{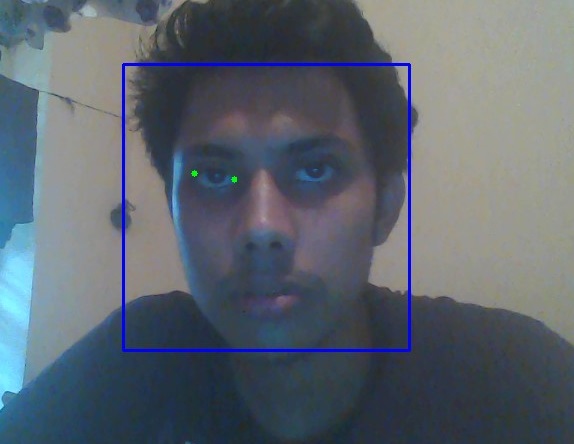}
\caption{Eye corner Detection}
\end{figure*}
\section{Santos and Proenca Method for Eye Corner Detection}
In the Santos and Proenca Method \cite{santos2011robust}, an eye image is used as an input to the detector. First, they obtain a noise-free iris binary segmentation mask from the eye image. The next step is segmentation of the sclera region. The sclera being the most unsaturated region in the eye image, an HSV transformation yields the lowest magnitudes on the saturation plane. The iris and sclera being segmented out, the next stage constitutes of the approximation of the eyelids contour. This is achieved using morphological dilation of the iris segmentation mask with a horizontal structuring element is carried out. This expands the iris regions horizontally. Finally, point-by-point multiplication between the dilated and the enhanced
data provides a good approximation to the eyelids contour. The subsequent step comprises of the generation of a set of candidate eye corner positions, which was performed using the Harris and Stephens corner detector.\\This method works well for an image having proper resolution and clarity. However, for inferior quality images, such as those obtained using a standard webcam, the performance of such methods is limited.
\section{Proposed Algorithm}
Our proposed scheme is a real-time camera-based system depicted as a schematic in Fig. \ref{fram}. For image acquisition, the scheme uses a standard webcam of resolution $640 \times 480$ pixels. The algorithm has been implemented in real-time using a Sony PS3 Eye webcam at 30 fps. First, face detection is carried out in a given frame, followed by eye detection. Eye detection is confined to a Region of Interest (ROI) in the detected face area. The RoI is selected based on our previous work \cite{dasgupta2013board}. This step finds each eye separately. The algorithm has been made robust using some preprocessing techniques for illumination correction. 
\subsection{Image Enhancement}
Preprocessing of the image is necessary for filtering out the noise and preserving relevant information. The filtering is particularly essential as the eye corner is sometimes not visible clearly because of illumination conditions. The pre-processing begins by equalizing the histogram of the image. We have used Contrast Limited Adaptive Histogram Equalization (CLAHE) technique \cite{singvi2012real} for this purpose. In this method, the image is fractioned into small blocks of size $8 \times 8$. Each of these blocks undergoes histogram equalization. This confines the histogram to a small region. An issue of this method is the amplification of noise if present. This issue is avoided by contrast limiting. After the CLAHE operation, bi-linear interpolation is applied to remove artifacts in tile borders.
\subsection{Face and Eye Detection}
The face and eye detection forms the first step in the algorithm for the localization of the eye corners. A classifier based on Haar-like features was selected for this stage. For optimal use, we have used parameters based on the earlier work \cite{gupta2011analysis}.
Once the face is detected, an ROI is selected from the facial region. The ROI selection scheme has been reported in the previous work \cite{dasgupta2013vision}. Now, the search for eyes is confined to a reduced area, which also reduces computation and improves the speed.
\subsection{Eye Corner Detection}
The corner detection operates on the detected eye region. 
\subsubsection{Sclera Segmentation}
As proposed in \cite{santos2011robust}, the sclera is segmented out by converting the RGB image to HSV and subsequently thresholding the saturation plane. For grayscale images, the color space conversion is not required, and the thresholding operation can be applied directly. The idea behind this lies in the fact that the sclera is the most unsaturated segment in the eye image. There remain certain noisy pixels because of the blood vessels in the sclera. This issue is addressed using the morphological opening of the sclera portion using an elliptical mask.
\subsubsection{Eyelid Contour Approximation} The sclera region is hence segmented out with the morphological post-processing. The boundary of the mask is overlaid on the eye image to obtain the eyelid contour approximation.
\subsubsection{Eye Corner Candidate selection}
The corner score is obtained by the sum of squared differences (SSD), $S(x,y)$ between the corresponding pixels of two patches in the eyelid contour image $I(x,y)$. The differential of the corner score is considered for finding out the corner candidates. A circular window  $w(u,v)$ is used, to make the response isotropic.
\begin{equation}
S(x,y)= \sum_{u} \sum_{v} w(u,v) (I(u+x,v+y)-I(u,v))^2
\end{equation}
With Taylor series expansion and proper approximation, we have 
\begin{equation}
S(x,y)=\begin{bmatrix}
x & y
\end{bmatrix}
H
\begin{bmatrix}
x\\ y
\end{bmatrix}
\end{equation}
The Harris matrix, $H$ is obtained as
\begin{equation}
H=\sum_{u} \sum_{v} w(u,v)\begin{bmatrix}
I_x^2 & I_xI_y\\ 
 I_xI_y & I_y^2
\end{bmatrix}
\end{equation}
\subsubsection{Post Pruning}
Since the actual eye corners will lie at the extreme ends of the eyelid contour, the eye corner pair having the farthest distance are selected as the eye corners. In cases of a tie, the mean corner is chosen as the correct corner estimate, among the corner candidates bearing equal distances.
\begin{figure}
\includegraphics[width=\linewidth]{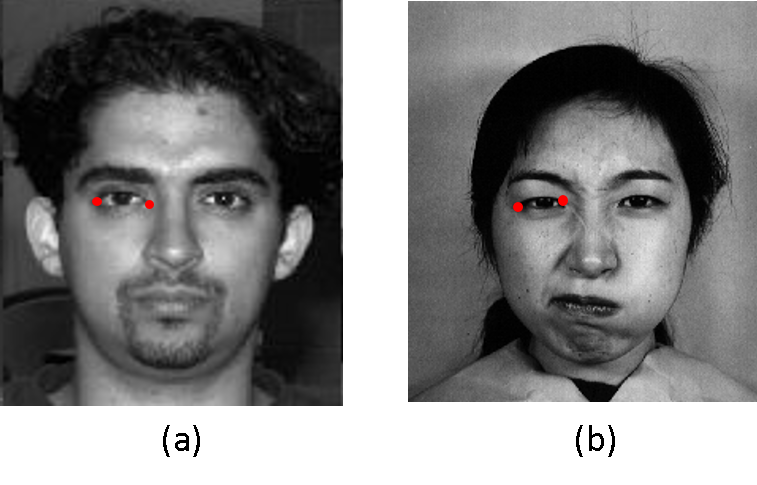}
\caption{Detection Results for images taken (a) Yale Database (b) JAFFE Database}
\end{figure}
\section{Results}
\begin{figure}
\includegraphics[width=\linewidth]{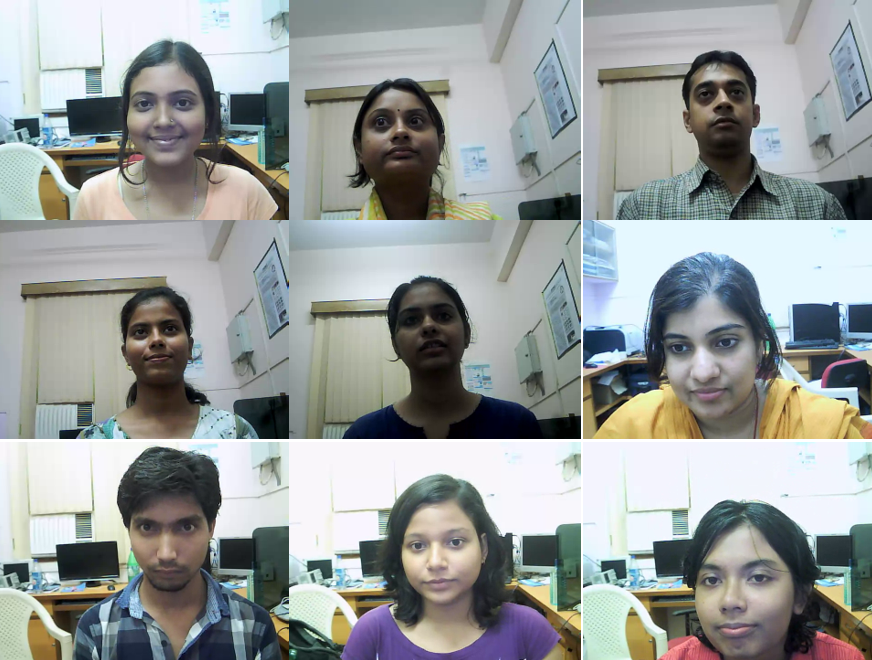}
\caption{Sample images from our created database}
\label{database}
\end{figure}

A face database of 30 subjects was prepared using a standard Sony PS3 web camera for testing the algorithm. Some sample images of the database is shown in Fig. \ref{database}. The algorithm has also been tested on  Yale Face Database \cite{lee2005acquiring} and JAFFE Database \cite{lyons1998japanese}. A sample set of 200 images total were randomly selected from the databases. The percentage of mean-squared pixel error in eye localization has been provided in Table \ref{results} for the different databases. The error is eye corner localization is based on the manual marking of the end-points. The processing-rate of the algorithm with online testing was found to be 16.2 fps while on offline database, it was 20.4 fps. The speed may be improved by using GPU based implementations, and employing parallel schemes.
\begin{table}[tbp]
\centering
\caption{Percentage Error in eye corner localization}
\label{results}
\begin{tabular}{|l|l|}
\hline
\textbf{Database}  & \textbf{\% Squared pixel error} \\ \hline
JAFFE database     & 4.5                             \\ \hline
Yale Face database & 6.5                             \\ \hline
IITKGP database    & 8.9                             \\ \hline
\end{tabular}
\end{table}
\section{Conclusion}
In this paper, we propose an algorithm that uses a standard web camera to localize effectively the eye corner. This is an improvisation over the Santos and Proenca Method. The significant modification lies in the post-pruning the eye corner candidates. The method has less than 10\% localization errors in all the three tested databases. A future scope of the work may be testing of the algorithm of infrared and near infra-red images \cite{happy2012video}, and make appropriate modifications so that the applications of the algorithm can be extended to areas where night vision is preferable.

\addtolength{\textheight}{-12cm}  
\section*{ACKNOWLEDGMENT}

The authors would like to acknowledge the subjects for participation in the experiment.


\bibliographystyle{IEEEtran}
\bibliography{library}
\end{document}